\documentclass[11pt,a4paper]{article}
\usepackage[hyperref]{emnlp-ijcnlp-2019}
\usepackage{times}
\usepackage{latexsym}

\usepackage{url}

\usepackage{algorithm}

\usepackage{algorithmic}
\usepackage{enumitem}
\usepackage{amsmath}
\usepackage{pifont}
\usepackage{array}
\usepackage{multirow}
\usepackage{graphicx}
\usepackage{subfig}
\usepackage{amssymb}

\usepackage{booktabs}

\usepackage{color, soul}
\usepackage{color}
\usepackage{colortbl}
\usepackage{tabularx}
\usepackage{tcolorbox}

\newtcbox{\mybox}[1][]
  {on line, arc = 0pt, outer arc = 0pt,
    colback = #1!15!white, colframe = #1!100!black,
    boxsep = 0pt, left = 1pt, right = 1pt, top = 2pt, bottom = 2pt,
    boxrule = 0pt, bottomrule = 1pt, toprule = 1pt}

\DeclareMathOperator{\LSTM}{LSTM}

\usepackage[justification=centering]{caption}
\newcommand{\tabincell}[2]{\begin{tabular}{@{}#1@{}}#2\end{tabular}}  

\aclfinalcopy 

\title{Human-Like Decision Making: Document-level Aspect Sentiment Classification via Hierarchical Reinforcement Learning}

\author{ Jingjing Wang$^{1}$, Changlong Sun$^{2}$, Shoushan Li$^{1,}$\thanks{\ \ Corresponding author}\ \ , Jiancheng Wang$^{1}$,\\ \textbf{Luo Si$^{2}$, Min Zhang$^{1}$, Xiaozhong Liu$^{2}$, Guodong Zhou$^{1}$}\\$^{1}$School of Computer Science and Technology, Soochow University, China\\$^{2}$Alibaba Group, China\\ \{djingwang, lishoushan, minzhang, gdzhou\}@suda.edu.cn,\\jiancheng.wang@qq.com, \{changlong.scl, xiaozhong.lxz, luo.si\}@alibaba-inc.com}

\date{}

\begin{document}
\maketitle
\begin{abstract}
Recently, neural networks have shown promising results on Document-level Aspect Sentiment Classification (DASC). However, these approaches often offer little transparency w.r.t. their inner working mechanisms and lack interpretability. In this paper, to simulating the steps of analyzing aspect sentiment in a document by human beings, we propose a new Hierarchical Reinforcement Learning (HRL) approach to DASC. This approach incorporates clause selection and word selection strategies to tackle the data noise problem in the task of DASC. First, a high-level policy is proposed to select aspect-relevant clauses and discard noisy clauses. Then, a low-level policy is proposed to select sentiment-relevant words and discard noisy words inside the selected clauses. Finally, a sentiment rating predictor is designed to provide reward signals to guide both clause and word selection. Experimental results demonstrate the impressive effectiveness of the proposed approach to DASC over the state-of-the-art baselines.
\end{abstract} 
\section{Introduction}
Document-level Aspect Sentiment Classification (DASC) is a fine-grained sentiment classification task in the field of sentiment analysis \cite{DBLP:journals/ftir/PangL07,DBLP:conf/acl/LiHZL10}. This task aims to predict the sentiment rating for each given aspect mentioned in a document-level review. For instance, Figure \ref{fig:one-example} shows a review document with four given aspects of a hotel (i.e., \emph{location}, \emph{room}, \emph{value}, \emph{service}). The goal of DASC is to predict the rating score towards each aspect by analyzing the whole document. In the last decade, this task has been drawing more and more interests of researchers in the Natural Language Processing community \cite{DBLP:conf/acl/TitovM08,DBLP:conf/emnlp/YinSZ17,DBLP:conf/coling/LiYZ18}. 
 \begin{figure}[t]
  \small
  \renewcommand\arraystretch{1}
 \scalebox{0.9}{
  \begin{tabular}{>{\small}p{1\columnwidth}}
  
				    \toprule
  \multicolumn{1}{c}{\textbf{Review Document}}\\
				    \toprule
$\pmb{\textcolor{red}{[}}$This hotel is \mybox[red]{close} to railway station$\pmb{ \textcolor{red}{]}}_{{\textcolor{red}{\textbf{Clause1}}}}$ $\pmb{\textcolor{red}{[ }}$and \mybox[red]{very convenient} to eat around$\pmb{\textcolor{red}{ ]}}_{{\textbf{\textcolor{red}{Clause2}}}}$ $\pmb{\textcolor{blue}{[ }}$but room of Hilton is \mybox[blue]{ a little uncomfortable}.\textbf{\textcolor{blue}{ ]}}$_{{\textbf{\textcolor{blue}{Clause3}}}}$ $\pmb{\textcolor{blue}{[}}$I'm often \mybox[blue]{nitpicking} for room decoration.$ \pmb{\textcolor{blue}{ ]}}_{{\textbf{\textcolor{blue}{Clause4}}}}$ $\pmb{\textcolor{black}{[}}$Besides, the price is \mybox[black]{very expensive}$\pmb{\textcolor{black}{\  ]}}_{{\textbf{\textcolor{black}{Clause5}}}}$ $\pmb{\textcolor{brown}{[}}$although the staff service is \mybox[brown]{professional}.$\pmb{\textcolor{brown}{]}}_{{\textbf{\textcolor{brown}{Clause6}}}}$\\
\bottomrule
 \multicolumn{1}{c}{\textbf{Rating of Each Aspect}}\\
				    \toprule

\hspace{0ex}\textbf{- \emph{location}:\hspace{0ex}} {\textcolor{red}{\ding {72} \ding {72} \ding {72} \ding {72} \ding {72}}} \ \hspace{0ex}(\textcolor{red}{\textbf{5}})
\hspace{3.9ex}\textbf{- \emph{room}:\hspace{2.2ex} } \textcolor{blue}{\ding {72} \ding {72} \ding {72}} \ding {73} \ding {73} \ \hspace{0ex}(\textcolor{blue}{\textbf{3}})\\

\hspace{0ex}\textbf{- \emph{value}:\hspace{2ex} }\textcolor{black}{\ding {72}} \ding {73} \ding {73} \ding {73} \ding {73} \ \hspace{0ex}(\textcolor{black}{\textbf{1}})
\hspace{3.6ex}\textbf{- \emph{service}:\hspace{0ex} }\hspace{0.5ex} \textcolor{brown}{\ding {72} \ding {72} \ding {72} \ding {72}} \ding {73} \  \hspace{0ex}(\textcolor{brown}{\textbf{4}})\\
\bottomrule
  \end{tabular}}

    \setlength{\belowcaptionskip}{-4 ex}
  \caption{An example of a review document, where clauses and words with different colors refer to different aspects.}
  \label{fig:one-example}
  \end{figure}
In previous studies, neural models have shown to be effective for performance improvement on DASC. Despite the advantages, these complex neural network approaches often offer little transparency w.r.t. their inner working mechanisms and suffer from the lack of interpretability. However, clearly understanding where and how such a model makes such a decision is rather important for developing real-world applications \cite{DBLP:journals/corr/abs-1807-02314,DBLP:journals/corr/abs-1801-00631}. 

As human beings, if asked to evaluate the sentiment rating for a specific aspect in a document, we often perform sentiment prediction in two steps. First, we select some \textbf{aspect-relevant snippets} (e.g., sentences/clauses) inside the document. Second, we select some \textbf{sentiment-relevant words} (e.g., sentiment words) inside these snippets to make a rating decision. For instance, for aspect \emph{location} in Figure \ref{fig:one-example}, we first select the aspect-relevant clauses, i.e., {Clause1} and {Clause2}, and then select sentiment-relevant words, i.e., ``\emph{ close}'' and ``\emph{very convenient}'' inside the two clauses, for making the rating decision (5 stars). 

Inspired by the above cognitive process of human beings, one ideal and interpretable solution for DASC is to select aspect-relevant clauses and sentiment-relevant words, discarding those noisy parts of a document for decision making. In this solution, two major challenges exist which are illustrated as follows.

The first challenge is how to select aspect-relevant clauses and discard those irrelevant and noisy clauses. For instance, for aspect \emph{location}, {{Clause5}} mentioning another aspect \emph{value} (only 1 star) may induce the noise and should be discarded, because the noise can provide wrong signals to mislead the model into assigning very low sentiment rating to aspect \emph{location}. One possible way to alleviate this noisy problem is to leverage the soft-attention mechanism as proposed in \newcite{DBLP:conf/coling/LiYZ18} and \newcite{DBLP:conf/ijcai/WangLLKZSZ18}. However, this soft-attention mechanism has the limitation that the \emph{softmax} function always assigns small but non-zero probabilities to noisy clauses, which will weaken the attention given to the few truly significant clauses for a particular aspect. Therefore, a well-behaved approach should discard noisy clauses for a specific aspect during model training.

The second challenge is how to select sentiment-relevant words and discard those irrelevant and noisy words. For instance, for aspect \emph{location}, words ``\emph{this}'', ``\emph{is}'' in {Clause1} are noisy words and should be discarded since they make no contribution to implying the sentiment rating. One possible way to alleviate this problem is to also leverage the soft-attention mechanism as proposed in \newcite{DBLP:conf/coling/LiYZ18}. However, this soft-attention mechanism may induce additional noise and lack interpretability because it tends to assign higher weights to some domain-specific words rather than real sentiment-relevant words \cite{mudinas2012combining,DBLP:conf/coling/ZouGZH18}. For instance, this soft-attention mechanism tends to regard the name of a hotel ``\emph{Hilton}'' with a good reputation in {Clause3} as a positive word which could mislead the model into assigning a higher rating to aspect \emph{room}. Therefore, a well-behaved approach should highlight sentiment-relevant words and discard noisy words for a specific aspect during model training.

{In this paper}, we propose a Hierarchical Reinforcement Learning (HRL) approach with a high-level policy and a low-level policy to address the above two challenges in DASC. First, a high-level policy is leveraged to select aspect-relevant clauses and discard noisy clauses during model training. Then, a low-level policy is leveraged to select sentiment-relevant words and discard noisy words inside the above selected clauses. Finally, a sentiment rating predictor is designed to provide reward signals to guide both clause and word selection. The empirical studies show that the proposed approach performs well by incorporating the clause selection and word selection strategies and significantly outperforms several state-of-the-art approaches including those with the soft-attention mechanism.
\section{Hierarchical Reinforcement Learning}

\begin{figure*}[t]
\centering
\vspace{-2ex}
\subfloat{
    \hspace{-1.8 ex}
 \includegraphics[scale=0.6]{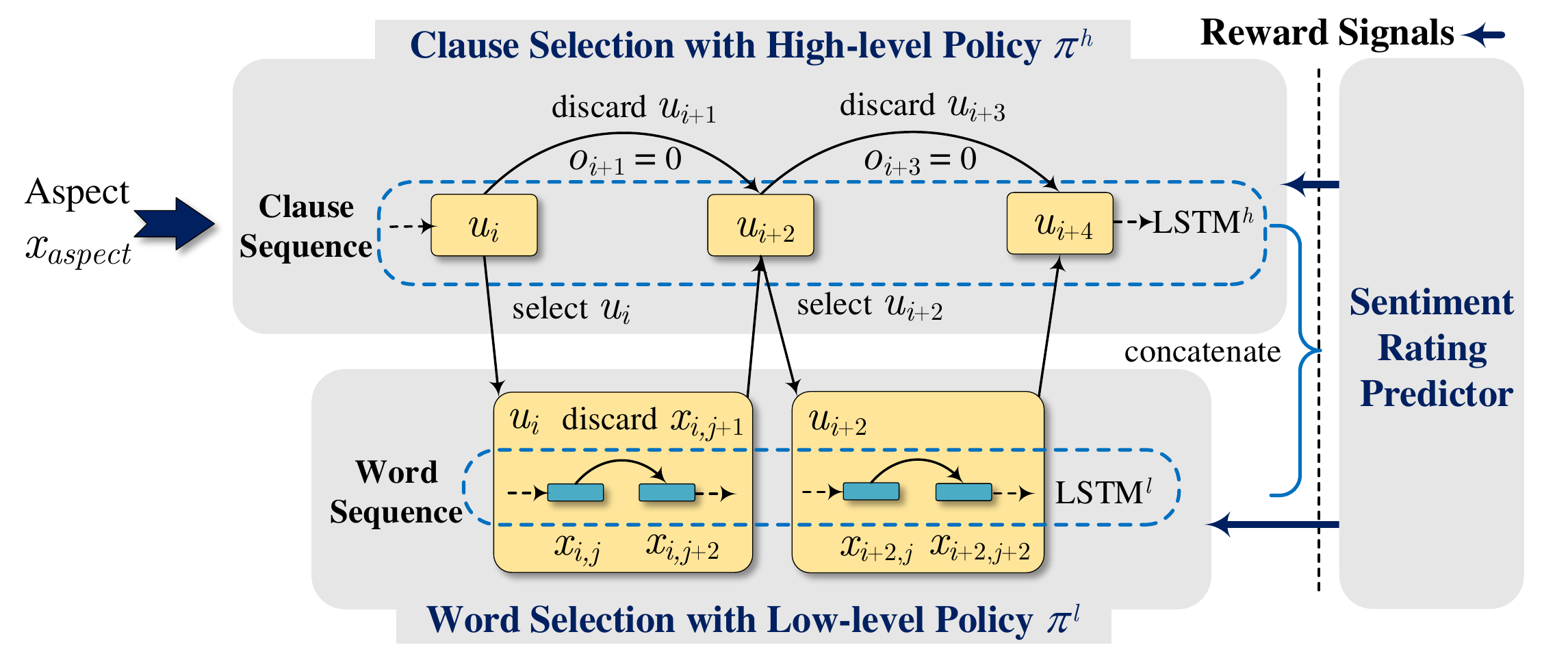}}

  \setlength{\belowcaptionskip}{-2 ex}
\caption{Overall architecture of our Hierarchical Reinforcement Learning (HRL) approach.}
\label{fig:approach}
\end{figure*}

Figure \ref{fig:approach} shows the overall framework of our Hierarchical Reinforcement Learning (HRL) approach which contains three components: a high-level policy for clause selection (Section 2.2); a low-level policy for word selection (Section 2.3); a sentiment rating predictor for providing reward signals to guide both the above clause and word selection (Section 2.4). 

As a preprocessing, we adopt RST style discourse segmentation\footnote{In preliminary experiments, we tried directly adopting sentence segmentation by following \newcite{DBLP:conf/coling/LiYZ18} rather than clause splitting and found that adopting clause splitting achieves better performance. Detailed comparison results are presented in ablation study of Section 3.2.} \cite{10014960635} to segment all documents in corpus $\mathcal{C}$ into Elementary Discourse Units (EDUs), and consider thse EDUs as clauses by following \newcite{DBLP:conf/ijcai/WangLLKZSZ18}.

In summary, we formulate the task of DASC as a semi-Markov Decision process \cite{DBLP:journals/ai/SuttonPS99}, i.e., hierarchical reinforcement learning with a high-level policy and a low-level policy. In particular, our HRL approach for DASC works as follows. Given a review document with a clause sequence and an aspect, the high-level policy decides whether a clause mentions this aspect. If yes, the high-level policy selects this clause and launches the low-level policy, which scans the words inside this selected clause one by one in order to select sentiment-relevant words. Otherwise, the high-level policy skips current clause and turns to the next clause until all clauses in the review document are scanned. During clause and word selection, a sentiment rating predictor is employed to provide reward signals to guide the above clause and word selection. 

\subsection{Clause Selection with High-level Policy}
Assume that a review document $\mathcal{D}$ with a given aspect ${x_{aspect}}$ has been segmented into a clause sequence $\{u_1,...,u_n\}$, high-level policy $\pi^h$ aims to select clause $u_i$ which truly mentions aspect ${x_{aspect}}$ and discard noisy ones. Here, clause $u_i$ consists of $k_i$ words $\{x_{i,1},...,x_{i,{k_i}}\}$.
Once a clause is selected, it is passed to the low-level policy for further word selection.

During clause selection, we adopt a stochastic policy as high-level policy $\pi^h$, which can generate a conditional probability distribution $\pi^h({o}|\cdot)$ over option (i.e., high-level action) sequence ${o}=[o_1,...,o_n]$. Here, $o_i=1$ indicates $u_i$ is selected; otherwise $o_i=0$ indicates $u_i$ is discarded. More specifically, we adopt a LSTM model $\LSTM^h$ to construct high-level policy $\pi^h$ for performing clause selection over the clause sequence. In $\LSTM^h$, the hidden state $\hat{v}_i \in \mathbb{R}^d$ of clause $u_i$ and memory cell $c^h_i$ at $i$-th time-step are given by,
\begin{equation}
\hat{v}_i,c^h_i=
\begin{cases}
f(\hat{v}_{i-1},c^h_{i-1},{v}_i)& o_{i}=1\\
\hat{v}_{i-1},c^h_{i-1} & o_{i}=0
\end{cases}
\end{equation}
where ${v}_i$ is the vector representation of clause $u_i$ and initialized by hidden state $\hat{w}_{i,k_i}$ of the last word $\hat{x}_{i,k_i}$ in clause $u_i$. Here, $\hat{w}_{i,k_i}$ is obtained from the pre-trained $\LSTM^l$ (presented in line 3 of Algorithm 1); $f$ denotes all gate functions and update function of $\LSTM^h$. Note that if $o_i=0$, $\LSTM^h$ will skip (i.e., discard) and not encode clause $u_i$, memory cell $c_i^h$ and hidden state $\hat{v}_i$ of current time-step $i$ are then directly copied from the previous time-step $i-1$.

In principle, the high-level policy $\pi^h$ uses a \textbf{Reward} to guide clause selection over the clause sequence. It samples an \textbf{Option} $o_i$ with the probability $\pi^h(o_i|s^h_i;\theta^h)$ at each \textbf{State} $s^h_i$. More concretely, the state, option and reward of $\pi^h$ are defined as follows.


 \textbf{$\bullet$ State.} 
The state $s^h_i$ at $i$-th time-step should provide adequate information for deciding to
select a clause or not for ${x_{aspect}}$. Thus, state $s^h_i \in \mathbb{R}^{4d}$ is composed of four parts, i.e., $\hat{v}_{i-1}$, $c^h_{i-1}$, $v_i$ and $v_{a}$, defined as $
s^h_i=\hat{v}_{i-1} \oplus c^h_{i-1} \oplus v_i \oplus v_{a}$, where $v_{a} \in \mathbb{R}^d$ is aspect embedding\footnote{Aspect embedding is initialized by an average of
its keywords embeddings. Keywords for aspects (e.g., keywords for aspect \emph{service} are service, smile etc.) are obtained from \newcite{DBLP:conf/emnlp/YinSZ17}.} of ${x_{aspect}}$. 

 \textbf{$\bullet$ Option.} $\pi^h$ samples \emph{option} $o_i \in \{0,1\}$ by the conditional probability $\pi^h(o_i|s^h_i;\theta^h)$ which could be cast as a binary classification problem. Thus, we adopt logistic function to define $\pi^h(o_i|s^h_i;\theta^h)$.
\vspace{-2ex}
\begin{equation}
\begin{split}
o_i \sim \pi^h(o_i|s^h_i;\theta^h)=o_i\sigma(W^hs^h_i+b^h)\\
+(1-o_i)(1-\sigma(W^hs^h_i+b^h))\hspace{2.5ex}
\end{split}
\end{equation}
where $\theta^h=\{W^h \in \mathbb{R}^{1 \times 4d}, b^h \in \mathbb{R}\}$ is the trainable parameter; $\sim$ denotes sampling operation; $\sigma$ denotes sigmod function.

 \textbf{$\bullet$ Reward.} In order to select aspect-relevant clauses inside a clause sequence $\{u_1,...,u_n\}$, given a sampled \emph{option} trajectory $\tau^h=(s^h_1,o_1,r^h_1,...,s^h_n,o_n,r^h_n)\sim \pi^h$, we compute the high-level cumulative reward $r_i^h$ at $i$-th time-step as follows:
 \begin{equation}
 \begin{split}
r^h_i=\lambda_1\sum_{t=i}^{n}\gamma^{t-i}\log \cos(v_{a}, \hat{v}_t)\hspace{11.2ex}\\+\hspace{0.4ex}\lambda_2\sum_{t=i}^n{\gamma^{t-i}r^l(u_t)}+\lambda_3\log p_{\theta}(y|\hat{v}_n)\hspace{3ex}
\end{split}
\end{equation}
where $r_i^h$ consists of three different terms: \textbf{1)} The first term $\log \cos(v_{a}, \hat{v}_t)$ is a {cosine intermediate reward} computed by $cosine$ similarity between aspect embedding $v_a \in \mathbb{R}^d$ and hidden state $\hat{v}_t \in \mathbb{R}^d$ of the $t$-th clause $u_t$. This reward provides aspect supervision signals to guide the policy to select aspect-relevant clauses. \textbf{2)} The second term $r^l(u_t)=\sum_{j=1}^{k_t}r^l_{t,j}$ is an intermediate reward from low-level policy after the word selection in the selected clause $u_t$ is finished. Note that if clause $u_t$ is discarded, $r^l(u_t)=0$. This reward provides a feedback to indicate how well clause selection is. \textbf{3)} The third term $\log p_{\theta}(y|\hat{v}_n)$ is a delay reward from sentiment rating predictor. After $\LSTM^h$ finishes all options, we feed the last hidden state $\hat{v}_n$ of $\LSTM^h$ to the softmax decoder of sentiment rating predictor and then obtain a rating probability $p_{\theta}(y|\hat{v}_n)$ for ground-truth rating label $y$ to compute this delay reward. This reward provides additional reward signals to guide policy to select discriminative clauses. Besides, $\gamma$ is the discount factor; $\lambda_1$, $\lambda_2$ and $\lambda_3$ are weight parameters.


\subsection{Word Selection with Low-level Policy}
Given a word sequence $\{x_{i,1},...,x_{i,{k_i}}\}$ of clause $u_i$ selected by the high-level policy, low-level policy $\pi^l$ aims to select the sentiment-relevant word $x_{i,j}$ and discard noisy ones.

During word selection, we still adopt a stochastic policy as low-level policy $\pi^l$, which can generate a conditional probability distribution $\pi^l({a}|\cdot)$ over action sequence ${a}=[a_{i,1},..., a_{i,k_i}]$, where $a_{i,j}=1$ indicates $j$-th word $x_{ij}$ in $i$-th clause is selected; $a_{i,j}=0$ indicates $x_{i,j}$ is discarded. Similar to clause selection, we adopt another LSTM model $\LSTM^l$ to construct low-level policy $\pi^l$ for performing word selection over word sequence $\{x_{i,1},...,x_{i,{k_i}}\}$ of each clause (Note that, as shown in Figure \ref{fig:approach}, $\LSTM^l$ is shared by all selected clauses from high-level policy). In $\LSTM^l$, the hidden state $\hat{w}_{i,j} \in \mathbb{R}^d$ of word $x_{i,j}$ and memory cell $c^l_{j}$ at $j$-th time-step (Here, we omit the clause index and only use $j$ to denote the $j$-th time-step in $i$-th clause $u_i$) are given by,
\begin{equation}
\hat{w}_{i,j},c^l_{i,j}=
\begin{cases}
f(\hat{w}_{i,j-1},c^l_{i,j-1},w_{i,j})  & a_{i,j}=1\\
\hat{w}_{i,j-1},c^l_{i,j-1} & a_{i,j}=0
\end{cases}
\end{equation}
where $w_{i,j}$ is word embedding of $x_{i,j}$. Similar to $\LSTM^h$, in $\LSTM^l$, if $a_{i,j}=0$, i.e., word $x_{i,j}$ is discarded, memory cell and hidden state of current time-step are directly copied from the previous time-step. Then, we illustrate state, action and reward of low-level policy as follows.


 \textbf{$\bullet$ State.} 
The state $s^l_{i,j}$ at $j$-th time-step should provide adequate information for deciding to select a word or not. Thus, the state $s^l_{i,j} \in \mathbb{R}^{3d}$ is composed of three parts, i.e., $\hat{w}_{i,{i-1}}$, $c^l_{i,{j-1}}$, $w_{i,j}$, defined as $
s^l_{i,j}=\hat{w}_{i,{j-1}} \oplus c^l_{i,{j-1}} \oplus w_{i,j}$.

 \textbf{$\bullet$ Action.} $\pi^l$ samples \emph{action} $a_{i,j} \in \{0,1\}$ by the conditional probability $\pi^l(a_{i,j}|s^l_{i,j};\theta^l)$. Thus, similar to high-level policy, we adopt logistic function to define $\pi^l(a_{i,j}|s^l_{i,j};\theta^l)$.
\begin{equation}
\begin{split}
a_{i,j} \sim \pi^l(a_{i,j}|s^l_{i,j};\theta^l)=a_{i,j}\sigma(W^ls^l_{i,j}+b^l)\\
+(1-a_{i,j})(1-\sigma(W^ls^l_{i,j}+b^l))\hspace{4.3ex}
\end{split}
\end{equation}
where $\theta^l=\{W^l \in \mathbb{R}^{1 \times 3d}, b^l \in \mathbb{R}\}$ is the trainable parameter.

 \textbf{$\bullet$ Reward.} Similarly, in order to select sentiment-relevant words inside a word sequence $\{x_{i,1},...,x_{i,{k_i}}\}$, given a sampled \emph{action} trajectory $\tau^l=(s^l_{i,1},a_{i,1},r^l_{i,1},...,s^l_{i,{k_i}},a_{i,{k_i}},r^l_{i,k_i})\sim \pi^l$,  we compute the low-level cumulative reward $r^l_{i,j}$ at $j$-th time-step as:
\begin{equation}
r_{i,j}^l=\lambda'_1\log p_\theta(y|\hat{w}_{i,k_i})+\lambda'_2 (-N')/N
\end{equation}
where $r_{i,j}^l$ consists of two terms: \textbf{1)} Similar to high-level policy, the first term $\log p_\theta(y|\hat{w}_{i,k_i})$ is a delay reward provided by sentiment rating predictor. After $\LSTM^l$ finishes all actions, we feed last hidden state $\hat{w}_{i,k_i}$ in $i$-th clause to softmax decoder of sentiment rating predictor and then we can obtain this delay reward. This reward provides rating supervision information to guide policy to select discriminative words, i.e., sentiment-relevant words. \textbf{2)} The second term $\gamma (-N')/N$ is a penalty delay reward. $N'= \sum_{j=1}^{k_i} a_{i,j}$ denotes the number of selected words. The basic idea of using this penalty reward is to select words as small as possible because sentiment-relevant words is usually a small subset of all words inside the clause. Note that, we could also adopt external sentiment lexicons to achieve this goal, but sentiment lexicons are difficult to obtain in many real-world applications. Besides, $\lambda'_1$, $\lambda'_2$ are weight parameters.

\subsection{Sentiment Rating Predictor}
The goal of sentiment rating predictor lies in two-fold. On one hand, during model training, the goal of sentiment rating predictor is to use a softmax decoder to provide rating probabilities as the reward signals (see Eq.(3) and Eq.(6)) to guide both clause and word selection. 

On the other hand, when model training is finished, i.e., both high-level and low-level policy finish all their selections, the goal of sentiment rating predictor is to perform DASC. Specifically, we first regard last state $\hat{v}_n$ of $\LSTM^h$ as the representation of all selected clauses while last state $\hat{w}_{n,k_n}$ of $\LSTM^l$ as the representation of all selected words. Then, we concatenate $\hat{v}_n$ and $\hat{w}_{n,k_n}$ to compute the final representation $z$ of the review document $\mathcal{D}$ as: $z=\hat{v}_n \oplus \hat{w}_{n,k_n}$,
where $\oplus$ denotes concatenation operation. Finally, to perform DASC, we feed $z$ to a softmax decoder as follows:

\textbf{Softmax Decoder.} We first feed $z$ to a softmax classifier
$m=Wz+b$, where $m \in \mathbb{R}^{C}$ is output vector; $\theta=\{W, b\}$ is trainable parameter. Then, the probability of labeling sentence with sentiment rating $\hat{y} \in [1,C]$ is computed by
$p_{\theta}(\hat{y}|z)=\frac{\exp(m_{\hat{y}})}{\sum_{\ell=1}^{C}\exp(m_{\ell})}$.
Finally, the label with the highest probability stands for the predicted sentiment rating for ${x_{aspect}}$.

Note that, for an aspect, if no clauses and words are finally selected from a review document, the model will assign a random rating for this aspect.

\subsection{Model Training via Policy Gradient and Back-Propagation}
The parameters in HRL are learned according to Algorithm \ref{alg:algorithm}. Specifically, these parameters can be divided into two groups: \textbf{1)} $\theta^h$ and $\theta^l$ of high-level policy $\pi^h$ and low-level policy $\pi^l$ respectively. \textbf{2)} $\theta$ of $\LSTM^h$, $\LSTM^l$ and softmax decoder. 

For $\theta^h$ of high-level policy, we optimize it with policy gradient (REINFORCE) \cite{DBLP:journals/ml/Williams92,DBLP:conf/nips/SuttonMSM99}. The policy gradient w.r.t. $\theta^h$ is computed by differentiating the maximized expected reward $J(\theta^h)$ as follows:
\vspace{-1ex}
\begin{align}
\nabla_{\theta^h} J(\theta^h)= \mathbb{E}_{{{\tau^h}} \sim \pi^h}[\sum_{i=1}^{n} \mathcal{R}^h\nabla_{\theta^h}\log \pi^h(o_i|s^h_i;\theta^h)]\hspace{0ex}
\end{align}
\vspace{-2ex}

\noindent where $\mathcal{R}^ h=r^h_i-b(\tau^h)$ is the advantage estimate of the high-level reward. Here, $b(\tau^h)$ is the \emph{baseline} \cite{DBLP:journals/ml/Williams92} which is used to reduce the variance of the high-level reward without altering its expectation theoretically. In practical use, we sample some trajectories $\tau^h_1,\tau^h_2,...,\tau^h_m$ over the clause sequence with the current high-level policy. The model will assign a reward score to each sequence according to the designed scores function, and then estimates $b(\tau^h)$ as the average of those rewards.
Similarly, the policy gradient w.r.t. $\theta^l$ of low-level policy is given by,
\vspace{-1ex}
\begin{align}
\nabla_{\theta^l} J(\theta^l)= \mathbb{E}_{\tau^l \sim \pi^l}[\sum_{j=1}^{k_i} \mathcal{R}^l\nabla_{\theta^l}\log \pi^l(a_{i,j}|s^l_{i,j};\theta^l)]
\end{align}
\vspace{-2ex}

\noindent where $\mathcal{R}^ l=r^l_{i,j}-b(\tau^l)$ is the advantage estimate of the low-level reward. Similarly, $b(\tau^l)$ is used to reduce the variance of the low-level reward.

For $\theta$, we optimize it with back-propagation. The objective of learning $\theta$ is to minimize the cross-entropy loss in the classification phase, i.e.,
\vspace{-0.5ex}
\begin{equation}
J(\theta)=\mathbb{E}_{(\mathcal{D},{x_{aspect}},y) \sim \mathcal{C}}[-\log p_\theta(y|z)]
+\frac{\delta}{2}||\theta||_{2}^{2}
\end{equation}
\vspace{-2ex}

\noindent where $(\mathcal{D}, {x_{aspect}}, y)$ denotes a review document $\mathcal{D}$ with a given aspect ${x_{aspect}}$ from corpus $\mathcal{C}$; $y$ is ground-truth sentiment rating for aspect ${x_{aspect}}$. $\delta$ is a $L_2$ regularization. 

\begin{algorithm}[tb]
\caption{Hierarchical reinforcement learning}
\label{alg:algorithm}
\small
\begin{algorithmic}[1] 
\STATE \textbf{Input:} Corpus $\mathcal{C}$; a review document $\mathcal{D}$ with a clause sequence $\{u_1,...,u_n\}$; a clause $u_i$ with a word sequence $\{x_{i,1},...,x_{i,{k_i}}\}$.

\STATE Initialize parameters $\theta$, $\theta^h$ and $\theta^l$ randomly;
\STATE Pre-train $\LSTM^l$ by forcing $\pi^l$ to select all words for classification, and update $\theta$ of $\LSTM^l$ by Eq.(9);
\STATE Pre-train $\LSTM^h$ by forcing $\pi^h$ to select all clauses for classification, and update $\theta$ of $\LSTM^h$ by Eq.(9);
\STATE Fix all parameters $\theta$ and update $\theta^h$, $\theta^l$ as follows:
\FOR{review document $\mathcal{D} \in \mathcal{C}$}
\FOR{clause $u_i\in \{u_1,...,u_n\}$}
\STATE Sample \emph{option} $o_i \sim  \pi^h(o_i|s^h_i;\theta^h)$;
\IF {option $o_i=1$}
\FOR{word $x_{i,j} \in \{x_{i,1},...,x_{i,{k_i}}\}$}
\STATE Sample \emph{action} $a_{i,j} \sim  \pi^h(a_{i,j}|s^l_{i,j};\theta^h)$;
\ENDFOR
\STATE Compute $r_{i,j}^l$ by Eq.(6);
\STATE Update $\theta^l$ by Eq.(8);
\ENDIF
\ENDFOR
\STATE Compute $r_i^h$ by Eq.(3);
\STATE Update $\theta^h$ by Eq.(7);
\ENDFOR
\end{algorithmic}
\end{algorithm}

\begin{table}[t]
\centering
\vspace{-1.5ex}
  \begin{tabular}{>{\small}l>{\small}c>{\small}>{\small}c>{\small}c>{\small}c}
  \toprule
  \textbf{Datasets} & \emph{\#documents}  & \emph{\#words/doc}   & \emph{\#Aspect}\\ 
  
  \hline
  {TripUser} & 58632 & 181.03 & 7 \\ 
{TripAdvisor} & 29391 & 251.7  & 7 \\ 
{BeerAdvocate}  & 51020 & 144.5 & 4  \\ 
  \toprule
  \end{tabular}
  \setlength{\belowcaptionskip}{-3 ex}
  \caption{Statistics of three datasets. \emph{\#words/doc} denotes the number of words (average per document). \emph{\#Aspect} denotes the number of aspects in each dataset. 
  The rating scales of TripAdvisor and BeerAdvocate are 1-5 and 1-10 respectively.}
  \label{tab:datasets} 
\end{table}

\setlength{\tabcolsep}{1.8pt}
\begin{table*}[t]
	\renewcommand{\arraystretch}{1.2}
	\addtolength{\tabcolsep}{1pt}
	\begin{center}
		\begin{small}
			\begin{tabular}{l |cc|cc|cc|cc|cc|cc}
				    \toprule
					  				 \multirow{3}{*}{\textbf{Approaches}} 	& \multicolumn{4}{c|}{\textbf{TripUser}} & \multicolumn{4}{c|}{\textbf{TripAdvisor}} &	\multicolumn{4}{c}{\textbf{BeerAdvocate}}\\ 
					   \cline{2-5}
					    \cline{6-9}
					   \cline{10-13}
					   
						&\multicolumn{2}{c|}{\emph{Development}} 
						&\multicolumn{2}{c|}{\emph{Test}} 
						&\multicolumn{2}{c|}{\emph{Development}}
						&\multicolumn{2}{c|}{\emph{Test}} 
						&\multicolumn{2}{c|}{\emph{Development}} 
						&\multicolumn{2}{c}{\emph{Test}}\\
					   
						   \cline{2-5}
					    \cline{6-9}
					   \cline{10-13}

				            &Acc.$\uparrow$ &MSE$\downarrow$
				            &Acc.$\uparrow$ &MSE$\downarrow$
				            &Acc.$\uparrow$ &MSE$\downarrow$
				            &Acc.$\uparrow$ &MSE$\downarrow$
				            &Acc.$\uparrow$ &MSE$\downarrow$
				            &Acc.$\uparrow$ &MSE$\downarrow$\\
				    \hline

					
					
				    
					{SVM} 
					&-  &- 
					&46.35$^\dagger$  & 1.025$^\dagger$
					
					&34.30$^\ddagger$  & 1.982$^\ddagger$  
					&35.26$^\ddagger$  & 1.963$^\ddagger$  
					
					&25.70$^\ddagger$ &3.286$^\ddagger$  
					&25.79$^\ddagger$ &3.270$^\ddagger$ \\

					{LSTM} 
					&53.23  &0.787  
					&52.74  &0.794    
					
					& 43.85$^\ddagger$ & 1.525$^\ddagger$ 
					& 44.02$^\ddagger$ & 1.470$^\ddagger$ 
					
					& 35.23$^\ddagger$ & 2.112$^\ddagger$ 
					& 34.78$^\ddagger$ & 2.097$^\ddagger$ \\
					
					
					
					MAMC
					&- &- 
					&55.49$^\dagger$  & 0.583$^\dagger$
					
					& 46.21$^\ddagger$ & 1.091$^\ddagger$ 
					& 46.56$^\ddagger$ & 1.083$^\ddagger$ 
					
					& 39.43$^\ddagger$ & 1.696$^\ddagger$ 
					& 38.06$^\ddagger$ & 1.755$^\ddagger$ \\

					HARN
					& - & - 
					& 58.15$^\dagger$ & 0.528$^\dagger$   
					
					& - & - 
					& 48.21$^\ddagger$ & 0.923$^\ddagger$
					
					&39.81  &1.672  
					&38.19  &1.751  \\
					
					HUARN            	
					& - & -
					& 60.70$^\dagger$    & 0.514$^\dagger$ 
					
					& - & -
					& - & - 
					
					& - & - 
					& - & - \\
					
					{C-HAN} 
					&58.49 & 0.602
					&57.38 & 0.543  
					
					& 47.61 & 0.914 
					& 47.08 & 0.955
					
					& 38.67 & 1.703 
					& 37.95 & 1.801 \\
					
					
					
					
					HS-LSTM             	
					&59.75  &0.566 
					&59.01  &0.524    
					
					&48.45  & 0.947 
					&46.84  & 1.013 
					
					&37.43  &1.870  
					&36.83  &1.912  \\

                    \hline

					RL-Word-Selection                	
					&60.15  &0.475 
					&59.55  &0.519    
					
					&48.55  &0.913 
					&48.51  & 0.917 
					
					&39.92  &1.648  
					&38.45  &1.697  \\
					
					RL-Clause-Selection                	
					&61.32  &0.433 
					&60.54  &0.461    
					
					&51.05  &0.762 
					&50.02  &0.781  
					
					&41.39  & 1.505 
					&39.76  & 1.622 \\

					{\textbf{HRL}} 
					& \textbf{62.97} & \textbf{0.336}
					& \textbf{62.84} & \textbf{0.351} 
					
					& \textbf{52.71} & \textbf{0.652}
					& \textbf{52.27} & \textbf{0.662} 
					
					& \textbf{43.41} & \textbf{1.416}
					& \textbf{41.39} & \textbf{1.503} \\
					\bottomrule
			\end{tabular}
             \setlength{\belowcaptionskip}{-4 ex}
				\caption{Comparison of our approaches and other baseline approaches to DASC. The results with symbol $\dagger$  are retrieved from \newcite{DBLP:conf/coling/LiYZ18} and those with $\ddagger$ are from \newcite{DBLP:conf/emnlp/YinSZ17}}

			\label{table:results1}
		\end{small}
	\end{center}
\end{table*} 

\section{Experimentation}
\subsection{Experimental Settings}
\textbf{Data.} We conduct our experiments on three public datasets on DASC, i.e., TripUser \cite{DBLP:conf/coling/LiYZ18}, TripAdvisor \cite{DBLP:conf/kdd/WangLZ10} and BeerAdvocate \cite{DBLP:conf/icdm/McAuleyLJ12,DBLP:conf/emnlp/LeiBJ16}. In the experiment, we adopt Discourse Segmentation Tool\footnote{http://alt.qcri.org/tools/discourse-parser/} to segment all reviews in the three datasets into EDUs (i.e., clauses). Moreover, we adopt training/development/testing settings (8:1:1) by following \newcite{DBLP:conf/emnlp/YinSZ17,DBLP:conf/coling/LiYZ18}. Table \ref{tab:datasets} shows the statistics of the three datasets.

\textbf{Implementation Details.} We adopt the pre-trained 200-dimension word embeddings provided by \newcite{DBLP:conf/emnlp/YinSZ17}. The dimension of LSTM hidden states is set to be 200. The other hyper-parameters are tuned according to the performance in the development set. Specifically, we adopt Adam optimizer \cite{DBLP:journals/corr/KingmaB14} with an initial learning rate of 0.012 for cross-entropy training and adopt SGD optimizer with a learning rate of 0.008 for all policy gradients training. For rewards of high-level and low-level policies, $\gamma$ is 0.8; $\lambda_1$, $\lambda_2$ and $\lambda_3$ are 0.25, 0.25 and 0.5 respectively. $\lambda'_1$, $\lambda'_2$ are 0.6 and 0.4. Additionally, the batch size is set to be 64, regularization weight is set to be $10^{-5}$ and the dropout rate is 0.2.

\textbf{Evaluation Metrics.} The performance is evaluated using {Accuracy} (Acc.) and {MSE} as \newcite{DBLP:conf/emnlp/YinSZ17}. Moreover, $t$-test is used to evaluate the significance of the performance difference between two approaches \cite{DBLP:conf/sigir/YangL99}.

\textbf{Baselines.} We compare HRL with the following baselines: \textbf{1) SVM} \cite{DBLP:conf/emnlp/YinSZ17}. This approach only adopts unigram, bigram as features to train an SVM classifier. \textbf{2) LSTM} \cite{DBLP:conf/emnlp/TangQL15}. This is a neural network approach to document-level sentiment classification which employs gated LSTM to learn text representation. \textbf{3) MAMC} \cite{DBLP:conf/emnlp/YinSZ17}. This approach employs hierarchical iterative attention to learn aspect-specific representation. This is a state-of-the-art approach to DASC. \textbf{4) HARN} \cite{DBLP:conf/coling/LiYZ18}. This approach adopts hierarchical attention to incorporate overall rating  and aspect information so as to learn aspect-specific representation. This is another state-of-the-art approach to DASC. \textbf{5) HUARN} \cite{DBLP:conf/coling/LiYZ18}. This approach extends HARN by integrating additional user information. This is another state-of-the-art approach to DASC. \textbf{6) C-HAN} \cite{DBLP:conf/ijcai/WangLLKZSZ18}. This approach adopts hierarchical attention to incorporate clause and aspect information so as to learn text representation. Although this is a state-of-the-art approach to sentence-level ASC, it could also be directly applied in DASC.
\textbf{7) HS-LSTM} \cite{DBLP:conf/aaai/ZhangHZ18}. This is a reinforcement learning approach to text classification, which employs a hierarchically LSTM to learn text representation. \textbf{8) RL-Word-Selection.}  Our approach which leverages only the word selection strategy by using the low-level policy. \textbf{9) RL-Clause-Selection.} Our approach which leverages only the clause selection strategy by using the high-level policy.

\setlength{\tabcolsep}{2.5pt}
\begin{table*}[t]
\vspace{-1ex}
	\renewcommand{\arraystretch}{1.2}
	\addtolength{\tabcolsep}{1pt}
	\begin{center}
		\begin{small}
			\begin{tabular}{l cc cc cc}
				    \toprule
\multirow{2}{*}{\textbf{Approaches}} & \multicolumn{2}{c}{\textbf{TripUser}} & \multicolumn{2}{c}{\textbf{TripAdvisor}} & \multicolumn{2}{c}{\textbf{BeerAdvocate}}\\ 

\cline{2-3}	
\cline{4-5}	
\cline{6-7}	

&Acc.$\uparrow$ &MSE$\downarrow$ 
&Acc.$\uparrow$ &MSE$\downarrow$
&Acc.$\uparrow$ &MSE$\downarrow$\\
     \hline
\textbf{HRL} & \textbf{62.84}& \textbf{0.351} & \textbf{52.27} & \textbf{0.662} & \textbf{{41.39}} & \textbf{{1.503}}\\
\hline
\hspace{1ex} {w/o}\hspace{1ex}\emph{cosine} intermediate  reward in Eq.(3)
&60.15& 0.487& 49.02& 0.811& 38.72  & 1.673  \\
					
\hspace{1ex} {w/o}\hspace{1ex}\tabincell{c}{penalty delay reward in Eq.(6)}
& 61.52& 0.425& 51.15& 0.711& 40.13  &1.577  \\
\hspace{1ex} {w/o}\hspace{1ex}\tabincell{c}{the representation of selected clauses}
& 58.40& 0.622&  46.33& 1.207 &38.61  &1.684  \\
\hspace{1ex} {w/o}\hspace{1ex}\tabincell{c}{the representation of selected words}
& 60.08& 0.497&  48.94& 0.886 &39.92  &1.626  \\
					
\hspace{1ex}\hspace{0ex} using sentence splitting instead of clause
& 59.74& 0.504& 50.11 & 0.842 &39.33  &1.651  \\
					
					\bottomrule
			\end{tabular}
			  \setlength{\belowcaptionskip}{-4 ex}

				\caption{Ablation study of HRL on three different datasets.}
				
			\label{table:results2}
		\end{small}
	\end{center}
\end{table*}

\subsection{Experimental Results}
Table \ref{table:results1} shows the performance comparison of different approaches. From this table, we can see that, all LSTM-based approaches outperform \textbf{SVM}, showing that LSTM has potentials in automatically learning text representations and can bring performance improvement for DASC.

Four state-of-the-art ASC approaches including \textbf{MAMC},  \textbf{HARN}, \textbf{HUARN} and \textbf{C-HAN} all perform better than \textbf{LSTM}. These results confirm the helpfulness of considering aspect information in DASC. Besides, we find the reinforcement learning based approach \textbf{HS-LSTM} without considering aspect information can achieve comparable performance with \textbf{MAMC}, \textbf{HARN}, \textbf{C-HAN}, and even beat \textbf{MAMC} on two datasets \textbf{TripUser} and \textbf{TripAdvisor}, which demonstrates that using reinforcement learning is a good choice to learn text representation for DASC.

Our approach \textbf{RL-Word-Selection} and \textbf{RL-Clause-Selection} outperform most above approaches and they only perform slightly worse than \textbf{HUARN}. This result encourages to perform clause or word selection in DASC. Among all these approaches, our approach \textbf{HRL} performs best and it significantly outperforms ($p$-value $<$ 0.01) strong baseline \textbf{HUARN} which actually considers some other kinds of external information, such as the overall rating and the user information. These results encourage to perform both clause and word selection in DASC.

\textbf{Ablation Study.}
Further, we conduct the ablation study of HRL to evaluate the contribution of each component. The results are shown in Table \ref{table:results2}. From this table, we can see that \textbf{1)} Using \emph{cosine} intermediate reward in Eq.(3) can averagely improve Acc. by 2.87\% on three different datasets. \textbf{2)} Using penalty delay reward in Eq.(6) can averagely improve Acc. by 1.23\%. \textbf{3)} To concatenate additional representation of the selected clauses in rating predictor can improve Acc. by 4.38\%. \textbf{4)} To concatenate additional representation of the selected words in rating predictor can improve Acc. by 2.72\%. \textbf{5)} Using the clause splitting instead of sentence splitting could improve Acc. by 2.44\%. This confirms that it is more appropriate to consider clauses as the segmentation units than sentences. This is because that 90\% of clauses contain only one opinion expression as proposed in \newcite{DBLP:journals/ldvf/BayoudhiGKB15}. For instance, as shown in Figure \ref{fig:one-example}, if we use the sentences as the segmentation units, Clause1-Clause3 will be assigned into one unit while they talk about two aspects, i.e., {\emph{location}} and {\emph{room}}. In this scenario, sentence selection will not be able to discard noisy parts inside the sentence for the aspect \emph{location} or \emph{room}.
\begin{figure}[t]
\centering

\subfloat{
    \hspace{-3 ex}
 \includegraphics[scale=0.66]{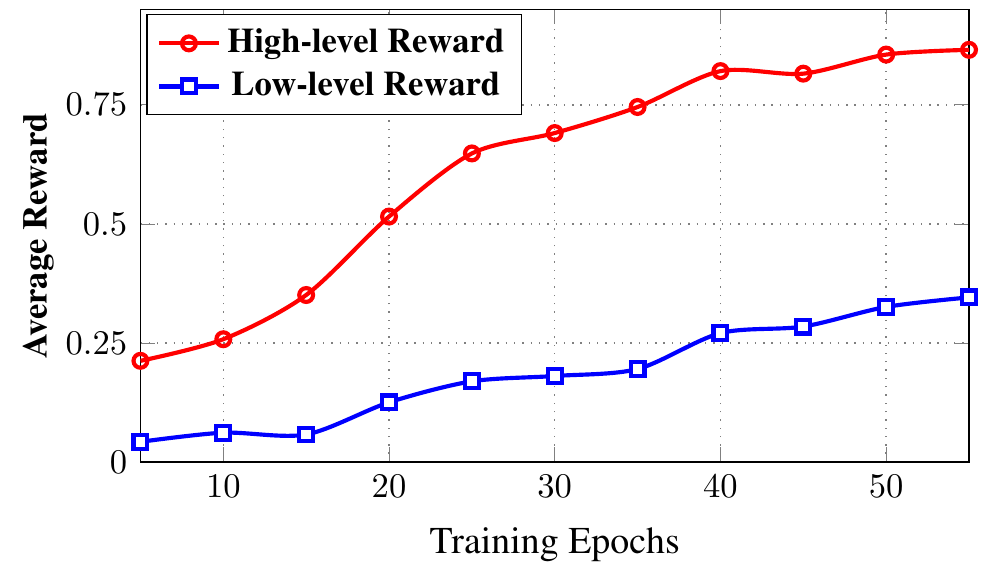}}
 \\
  \setlength{\belowcaptionskip}{-2 ex}
\caption{Training reward with different epochs}
\label{fig:reward}
\end{figure}

\begin{figure*}[t]
\centering
   \vspace{-1.5 ex}
\subfloat[For aspect \emph{\textbf{location}}]{
 \includegraphics[scale=1.3]{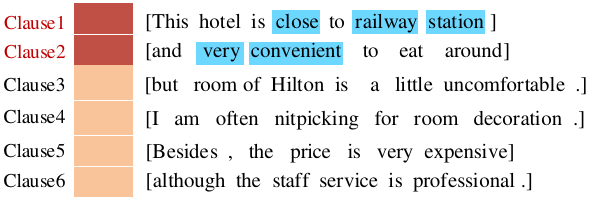}}
 \subfloat[For aspect \emph{\textbf{room}}]{
 \includegraphics[scale=1.3]{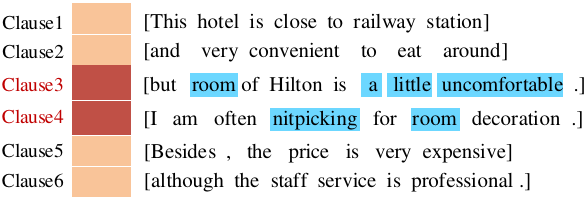}}
 \\
  \setlength{\belowcaptionskip}{-2 ex}
\caption{Visualizations of clause selection (along with the row $\downarrow$) and word selection (along with the column $\rightarrow$) for aspect (a) \emph{\textbf{location}} and (b) \emph{\textbf{room}}. Red denotes the clause has been selected, blue denotes the word has been selected and other colors denote the token has been discarded.}
\label{fig:visual}
\end{figure*}

\section{Analysis and Discussion}

\textbf{Analysis of HRL Training.} Figure \ref{fig:reward} shows two average rewards (each epoch) of high-level and low-level policy on BeerAdvocate respectively. To clearly observe the change of the reward, following \newcite{DBLP:journals/corr/LillicrapHPHETS15}, all rewards are normalized to $(0,1)$. From this figure, we can see that, both the high-level and low-level reward increase as the training algorithm iterates. This result indicates that our HRL approach is capable of stably revising its policies to obtain more discriminative clauses and words for better performing.

\textbf{Analysis of Clause and Word Selection.} Figure \ref{fig:visual} shows visualizations of our HRL approach which performs the clause selection and word selection on a review document. From this figure, we can see that HRL is able to precisely select aspect-relevant clauses, i.e., Clause1 and Clause2, for aspect {\emph{location}} while select Clause3 and Clause4 for {\emph{room}}. Further, HRL is able to select all sentiment-relevant words, such as ``\emph{close}'' and ``\emph{very convenient}'' for aspect {\emph{location}}, while ``\emph{a little uncomfortable}'' and ``\emph{nitpicking}''for {\emph{room}}.

\textbf{Error Type Breakdown.} We analyze error cases in the experiments and broadly categorize them into three types: \textbf{(1)} The first type of errors are due to negation words. For instance, for the review ``\emph{The taste of this beer is not good, don't buy it}'', HRL could precisely select the sentiment ``\emph{good}'', but fail to select the negation word  ``\emph{not}''. This inspires us to work on optimizing our approach in order to capture negation scope better in our future work. \textbf{(2)} The second type of errors are due to comparative opinions. For instance, for the review ``\emph{The room of Sheraton is much better than this one.}'', HRL incorrectly predicts high rating (5 stars) to aspect \emph{room}. It would be interesting to see if incorporating syntactic information can solve this problem and bring performance improvement. \textbf{(3)} Finally, some errors are due to mistakes during clause splitting (i.e., EDU splitting). For instance, for the review ``[\emph{This hotel having good location}] [\emph{often needs lots of time to check in.}]'', it is assigned into one clause while it talks two aspects, i.e., \emph{location} and \emph{check in/front desk}. This encourages to improve the performance of clause splitting for informal review texts.

\section{Related Work}
\textbf{Aspect Sentiment Classification.} Traditional studies for DASC mainly focus on feature engineering to explore efficient features for DASC \cite{DBLP:conf/acl/TitovM08,DBLP:conf/icdm/LuOCT11,DBLP:conf/icdm/McAuleyLJ12}.
Recently, neural networks with the characteristic of automatically mining features have shown promising results on DASC. \newcite{DBLP:conf/emnlp/LeiBJ16} focused on extracting rationales for aspects and build a neural text regressor to predict aspect rating; \newcite{DBLP:conf/emnlp/YinSZ17} focused on using hierarchical iterative attention to learn aspect-specific text representation for DASC; \newcite{DBLP:conf/coling/LiYZ18} employed a hierarchical attention approach to DASC which incorporates both the external user and overall rating information. Besides, neural networks have been widely adopted for performing a closely related task, i.e., Sentence-level Aspect Sentiment Classification \cite{DBLP:conf/emnlp/WangHZZ16,DBLP:conf/emnlp/TangQL16,DBLP:conf/aaai/WangL18a}.

\textbf{Reinforcement Learning.} In recent years, reinforcement learning has been applied successfully to some NLP tasks. \newcite{DBLP:journals/corr/Guo15b} employed deep Q-learning to improve the seq2seq model for the text generation task; \newcite{DBLP:journals/corr/LiMRGGJ16} showed how to apply deep reinforcement learning to model future reward in the chatbot dialogue task; \newcite{DBLP:journals/corr/abs-1811-03925} employed hierarchical reinforcement learning to model the relation extraction task; \newcite{DBLP:conf/aaai/ZhangHZ18} combined LSTM with reinforcement learning to learn structured representations for the text classification task, which is inspirational to our approach.

Unlike all above studies, inspired by the cognitive process of human beings, this paper proposes a new HRL approach to DASC task. To the best of our knowledge, this is the first attempt to address DASC with HRL. 


\section{Conclusion}
In this paper, we propose a hierarchical reinforcement learning approach to DASC. The main idea of the proposed approach is to perform sentiment classification like human beings. Specifically, our approach employs a high-level policy and a low-level policy to perform clause selection and word selection in DASC respectively. Experimentation shows that both the clause and word selection are effective for DASC and the proposed approach significantly outperforms several state-of-the-art baselines for DASC.

In our future work, we would like to solve other challenges in DASC, e.g., negation detection problem, to further improve the performance. Furthermore, we would like to apply our HRL approach to other sentiment analysis tasks, such as aspect and opinion co-extraction, and dialog-level sentiment analysis.

\section*{Acknowledgments}
We thank our anonymous reviewers for their helpful comments. This work was supported by three NSFC grants, i.e., No.61672366, No.61702149 and No.61525205. This work was also supported by the joint research project of Alibaba Group and Soochow University, and a Project Funded by the Priority Academic Program Development of Jiangsu Higher Education Institutions (PAPD).

\bibliography{emnlp-ijcnlp-2019}
\bibliographystyle{acl_natbib}

\end{document}